\documentclass[conference]{IEEEtran}
\IEEEoverridecommandlockouts
\usepackage{cite}
\usepackage{amsmath,amssymb,amsfonts}
\usepackage{algorithmic}
\usepackage{graphicx}
\usepackage{textcomp}
\usepackage{tabularx}
\usepackage{xcolor}
\usepackage{footnote}  
\usepackage{makecell}
\usepackage{bm}
\usepackage{booktabs}
\usepackage{hyperref}
\def\BibTeX{{\rm B\kern-.05em{\sc i\kern-.025em b}\kern-.08em
    T\kern-.1667em\lower.7ex\hbox{E}\kern-.125emX}}
\begin{document}

\title{3D-Telepathy: Reconstructing 3D Objects from
EEG Signals
\thanks{\\$\dag$ The two authors contribute equally to this work.\\
$*$ Corresponding authors. gespring@hdu.edu.cn, cmwangalbert@gmail.com}
}

\author{\IEEEauthorblockN{1\textsuperscript{st} Yuxiang Ge$^{\dag}$}
\IEEEauthorblockA{\textit{Hangzhou Dianzi University}\\
Hangzhou, China \\
}\\
\IEEEauthorblockN{3\textsuperscript{rd} Zhaojie Fang}
\IEEEauthorblockA{\textit{Hangzhou Dianzi University}\\
Hangzhou, China \\
}\\

\IEEEauthorblockN{6\textsuperscript{th} Nannan Li}
\IEEEauthorblockA{\textit{Macau University of Science and Technology} \\
Macau, China \\
}
\and
\IEEEauthorblockN{1\textsuperscript{st} Jionghao Cheng$^{\dag}$}
\IEEEauthorblockA{\textit{Hangzhou Dianzi University}\\
Hangzhou, China \\
}\\
\IEEEauthorblockN{4\textsuperscript{th} Gangyong Jia}
\IEEEauthorblockA{\textit{Hangzhou Dianzi University}\\
Hangzhou, China \\
}\\

\IEEEauthorblockN{7\textsuperscript{th} Ahmed Elazab}
\IEEEauthorblockA{\textit{Shenzhen University}\\
Shenzhen, China \\
}
\and
\IEEEauthorblockN{2\textsuperscript{nd} Ruiquan Ge$^{*}$}
\IEEEauthorblockA{\textit{Hangzhou Dianzi University}\\
Hangzhou, China \\
}\\


\IEEEauthorblockN{5\textsuperscript{th} Xiang Wan}
\IEEEauthorblockA{\textit{Shenzhen Research Institute of Big Data}\\
Shenzhen, China \\
}\\


\IEEEauthorblockN{8\textsuperscript{th} Changmiao Wang$^{*}$}
\IEEEauthorblockA{\textit{Shenzhen Research Institute of Big Data}\\
Shenzhen, China \\
}
}

\maketitle

\begin{abstract}

Reconstructing 3D visual stimuli from Electroencephalography (EEG) data holds significant potential for applications in Brain-Computer Interfaces (BCIs) and aiding individuals with communication disorders. Traditionally, efforts have focused on converting brain activity into 2D images, neglecting the translation of EEG data into 3D objects. This limitation is noteworthy, as the human brain inherently processes three-dimensional spatial information regardless of whether observing 2D images or the real world. The neural activities captured by EEG contain rich spatial information that is inevitably lost when reconstructing only 2D images, thus limiting its practical applications in BCI. The transition from EEG data to 3D object reconstruction faces considerable obstacles. These include the presence of extensive noise within EEG signals and a scarcity of datasets that include both EEG and 3D information, which complicates the extraction process of 3D visual data. Addressing this challenging task, we propose an innovative EEG encoder architecture that integrates a dual self-attention mechanism. We use a hybrid training strategy to train the EEG Encoder, which includes cross-attention, contrastive learning, and self-supervised learning techniques. Additionally, by employing stable diffusion as a prior distribution and utilizing Variational Score Distillation to train a neural radiation field, we successfully generate 3D objects with similar content and structure from EEG data.
Our code is available at \href{https://github.com/gegen666/EEGTo3D}{https://github.com/gegen666/EEGTo3D}.
\end{abstract}


\section{Introduction}
In our daily lives, our brains process spatial information from both two-dimensional images and the three-dimensional real world\cite{b1}. This ability is crucial for perceiving objects in terms of distance and depth\cite{b1}. Reconstructing 3D visual objects from Electroencephalography (EEG) data offers exciting possibilities as well as significant challenges in the field of brain-computer interfaces. Recent studies\cite{b2,b3} have successfully employed Diffusion\cite{b5} model-based techniques to reconstruct high-quality 2D images from EEG signals. Song \textit{et al.}\cite{b38} achieved notable 2D reconstruction results by using contrastive learning to align EEG embeddings with image embeddings. Li \textit{et al.}\cite{b29} further enhanced EEG-to-2D image generation by introducing a novel EEG encoder architecture and implementing a two-stage image generation strategy.

\begin{figure}[htbp]
    \centering
    \includegraphics[width=1\linewidth]{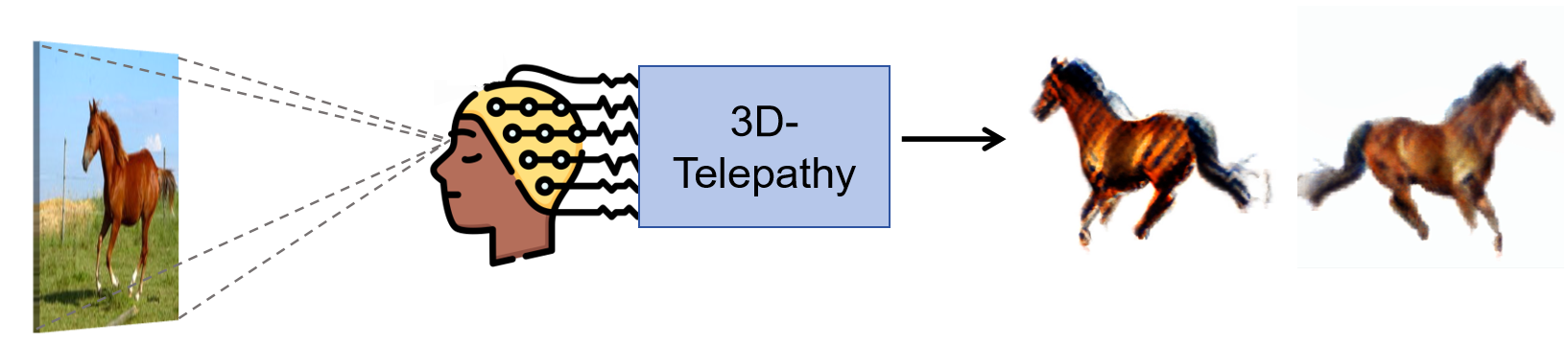}
    \caption{Schematic of the basic workflow of 3D-Telepathy.}
    \label{introduce_workflow}
\end{figure}

Our research aims to model this 3D visual capability of the brain. The central objective is to reconstruct 3D objects from EEG signals. This task involves extracting EEG's visual features and integrating its spatial and structural dimensions. Existing research on translating brain activity to 2D images\cite{b8,b9,b10,b29,b38} and multimodal 3D generation\cite{b11,b12,b13} offers a substantial theoretical foundation. Building on this foundation, we introduce an end-to-end EEG-3D model called 3D-Telepathy for converting EEG signals into high-quality 3D visual objects, as illustrated in Fig.~\ref{introduce_workflow}. The model's general structure involves inputting EEG signals to guide Neural Radiation Field (NeRF)\cite{b14} in obtaining the corresponding 3D representation.


Fig.~\ref{fig:model_frame} provides an overview of our model. In the first stage, the model processes EEG signals through an EEG encoder to derive feature vectors. This stage focuses on extracting both semantic and spatial information from EEG data, utilizing an EEG-image dataset\cite{b15} trained with a cross-attention, self-supervised contrastive learning framework. To ensure the accuracy of the feature vectors, we propose an EEG encoder architecture that incorporates a dual self-attention mechanism. This design effectively extracts and integrates the spatiotemporal information embedded in EEG signals.

The second stage of the model involves taking the EEG feature vectors obtained from the first stage and using them to generate 3D visual information via a model employing NeRF as the medium for 3D visualization. The goal here is to train a module guided by the EEG feature vectors to accurately produce the corresponding 3D visual object. In this stage, we employ the Variational Score Diffusion (VSD) method to train NeRF, enabling precise 3D representation. Our contributions include the following:

$\bullet$ We developed a brain decoding framework that can decode 3D visual information from EEG signals in an end-to-end manner, achieving high-fidelity reconstruction of 3D objects. This advancement further extends the possibilities of cross-modal 3D generation.

$\bullet$ Our proposed EEG encoder, along with the cross-attention self-supervised contrastive learning framework, effectively extracts high-level visual features from EEG data. These features are sufficient to support the generation of 3D objects.

$\bullet$ We introduced an EEG encoder architecture that incorporates a dual self-attention mechanism. This design efficiently extracts and integrates the spatiotemporal information embedded in EEG signals.

\section{Related Work}

\subsection{Neural Signals for Visual Decoding}
Decoding visual information from the brain has been a crucial research objective in neuroscience and computer science \cite{b39}. While significant progress has been made in decoding two-dimensional visual stimuli \cite{b8,b29}, accurately reconstructing the shape and spatial information of three-dimensional objects remains a major challenge. Although fMRI has been applied to reconstruct three-dimensional visual stimuli \cite{b13}, the expensive equipment and low temporal resolution of fMRI make it difficult to meet the demands of real-time brain-computer interface applications. In contrast, EEG emerges as a more promising option due to its high temporal resolution and portability. Moreover, existing methods primarily rely on limited contrastive supervised learning \cite{b29,b38}, failing to fully utilize the intrinsic correlation between EEG signals and visual stimuli.

\subsection{Neural Decoding Using EEG Data}
Previous studies have demonstrated the advantages of temporal-spatial feature extraction modules in processing neural signal data. For instance, the work of Li \textit{et al.} \cite{b29} and Song \textit{et al.} \cite{b38} has achieved notable results in EEG visual decoding tasks. Through the introduction of Cross Attention and self-supervised learning strategies, researchers have found that they can enhance the Encoder's ability to represent neural signal features \cite{b10}. Additionally, methods by Yu \textit{et al.}\cite{b8} have further improved the quality of visual reconstruction by incorporating latent diffusion models \cite{b5}. These advances provide an important technical foundation for reconstructing three-dimensional visual information from EEG signals.

\subsection{Cross-modal 3D Generation}
In recent years, cross-modal 3D generation technology, particularly Text-to-3D generation methods, has made remarkable progress. These methods can be primarily categorized into two types: NeRF-based methods and explicit 3D representation methods. DreamFusion \cite{b12} proposed a breakthrough framework that trains NeRF using pretrained 2D diffusion models to generate 3D content. ProlificDreamer \cite{b11} improved upon DreamFusion by enhancing the geometric details and surface textures of generated objects while accelerating convergence. Recent research such as Zero123 \cite{b40} further explores generating view-consistent 3D content from single-view images. These technologies provide important technical references for reconstructing 3D visual content from EEG signals. In particular, the innovations of these methods in handling view continuity and geometric consistency offer valuable insights for addressing key challenges in EEG-to-3D reconstruction tasks. However, compared to text-to-3D generation, reconstructing 3D objects from EEG signals faces greater challenges, primarily due to the high-noise characteristics of EEG signals and the complexity of cross-modal mapping.

\section{Methods}

\begin{figure*}[t!]
    \centering
    \includegraphics[width=0.9\linewidth]{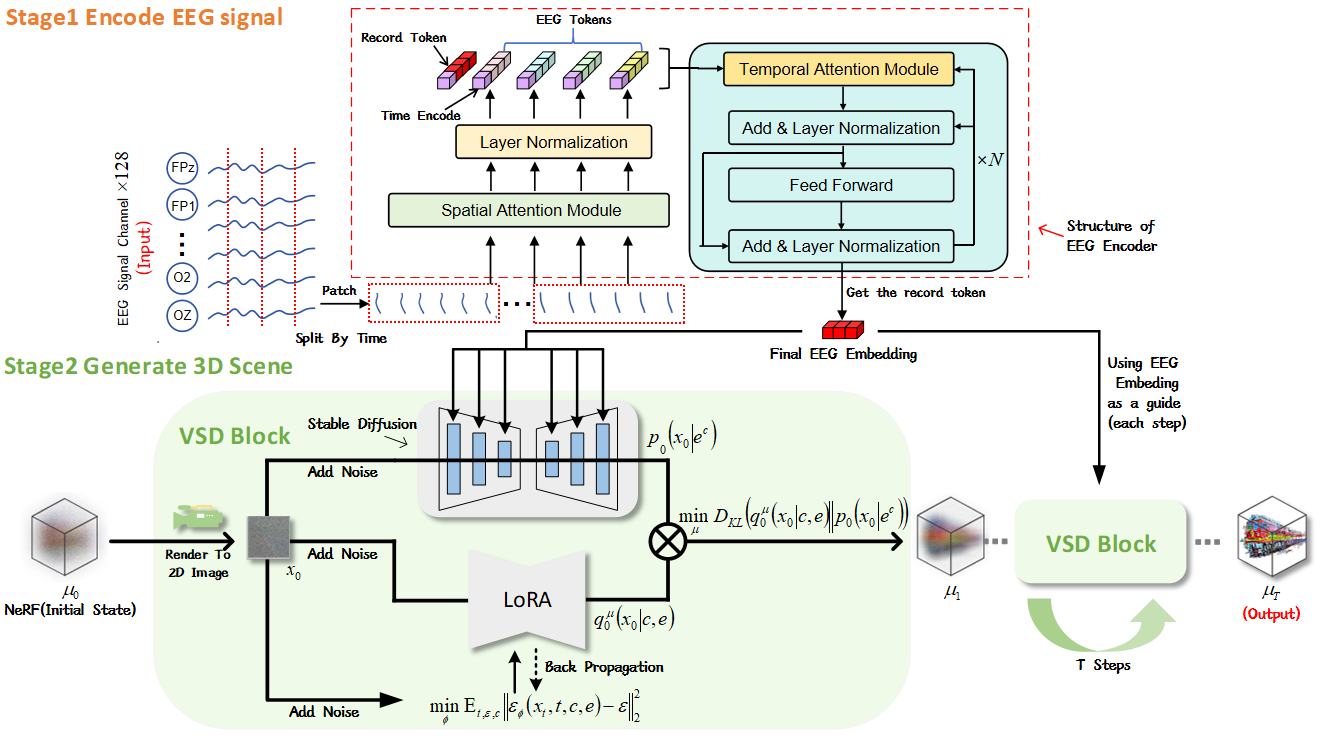}
    \caption{Overview of the 3D-Telepathy Framework. In Stage 1, we propose an EEG encoder architecture featuring a dual self-attention mechanism. The first layer of self-attention captures spatial dependencies among different electrode positions in EEG signals, while the second layer models captures long-range dependencies between temporal sequence features. Additionally, we introduce a record token that integrates relationships and features from all EEG tokens during the training process. In Stage 2, the 3D generation component utilizes the EEG embeddings from the first stage as conditioning signals, mapping them to the U-Net modules. Through the VSD method for NeRF training, our framework ultimately generates three-dimensional visual objects.}
    \label{fig:model_frame}
\end{figure*}

The architecture of our method, depicted in Fig.~\ref{fig:model_frame}, demonstrates an end-to-end framework for generating 3D objects directly from EEG signals. This approach comprises two primary phases. In the first phase, the EEG encoder extracts visual stimulus information from the EEG signals and transfers this information as embeddings to each layer of the U-Net in the Stable Diffusion model. The second phase (explained in Section \ref{s3}) involves training a NeRF using the VSD training framework, which utilizes the mapped embeddings. This results in a 3D object that embodies the visual stimulus information encoded in the EEG signals. Detailed explanations of the EEG encoder's internal structure and training strategy are provided in Section \ref{s1} and Section \ref{s2}.

\subsection{EEG Encoder}\label{s1}
We introduce an innovative encoder architecture designed to transform raw EEG signals into a feature representation space, as illustrated in Fig.~\ref{fig:model_frame}. This model processes EEG signals recorded during image viewing by segmenting the input data into multiple groups. Each group contains signals from all channels but spans different time periods. Our approach distinguishes itself from existing EEG encoder architectures by employing a hierarchical dual self-attention mechanism. The first layer of self-attention captures spatial dependencies among different electrode positions, converting the EEG signals into a set of uniformly formatted and interconnected EEG tokens. The second layer focuses on modeling long-range dependencies within the temporal sequence features. Prior to this second layer, we introduce a record token that captures correlations among EEG tokens during the self-attention process and incorporates their intrinsic feature information. Consequently, the record token effectively serves as the final EEG embedding, aggregating information from all EEG tokens to produce the encoder's output.

\subsection{EEG Encoder Training Strategy}\label{s2}
Numerous studies have highlighted the critical importance of high-quality EEG encoders in accurately reconstructing visual stimuli. However, current mainstream methods are largely confined to establishing connections between EEG signals and visual stimuli using contrastive learning techniques \cite{b29,b38}. Although these contrastive learning-based training strategies have proven effective, we contend that relying solely on a single contrastive learning approach may not fully harness the potential of EEG encoders. 

\begin{figure}[htp]
    \centerline{\includegraphics[width=1\linewidth]{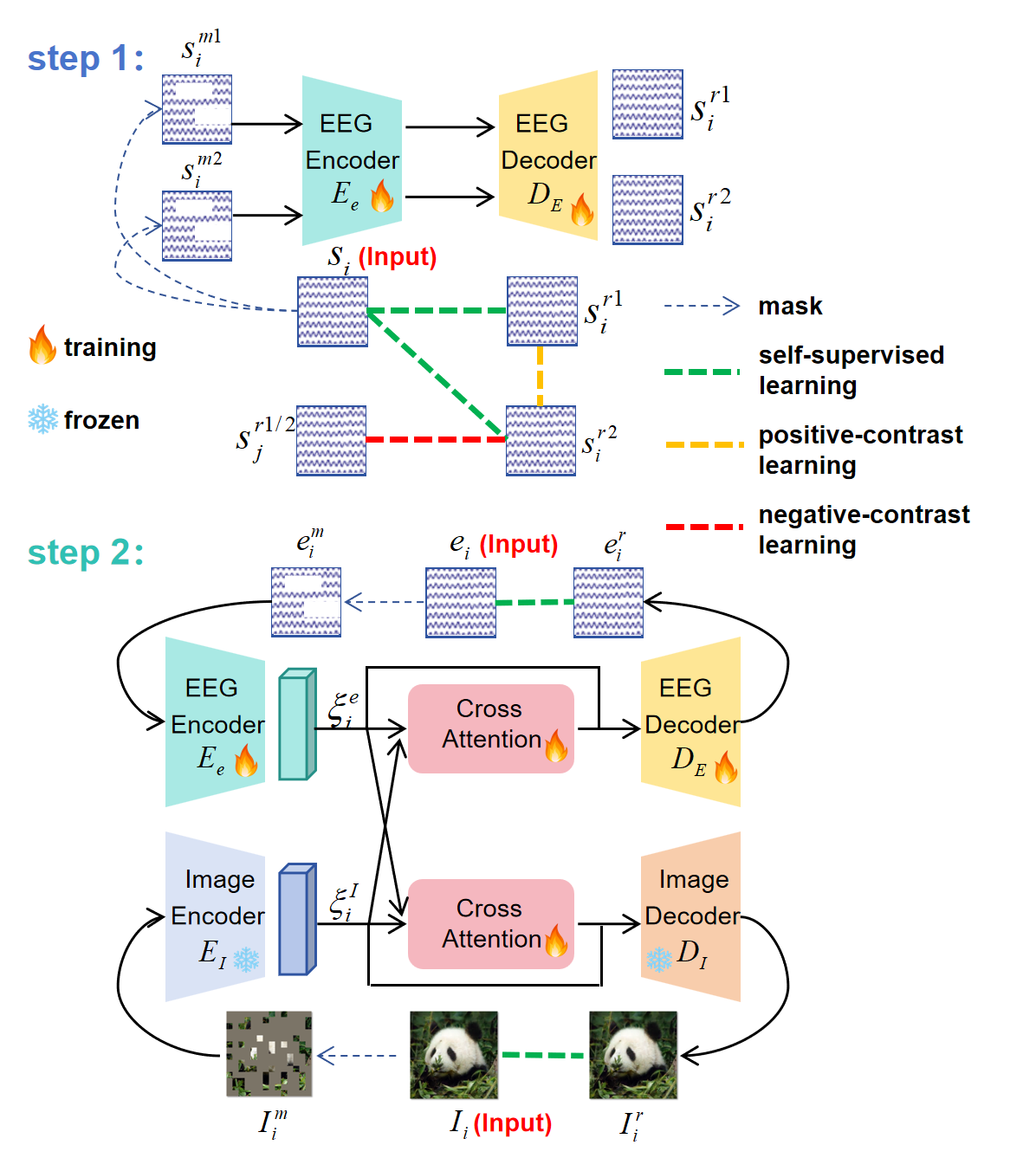}}
    \caption{Our proposed EEG encoder training strategy. In the first step, the encoder learns intrinsic feature representations of EEG signals through self-supervised learning, enhancing its ability to discriminate between different EEG signal patterns. In the second step, we introduce a cross-attention module to establish associations between EEG signals and their corresponding visual stimuli, thereby improving the encoder's capability to capture visual semantic information within EEG signals. $s_i$ and $e_i$ are the real EEG signal. $s_j^{r_{1/2}}$ represents the reconstruction results of other EEG signals, $\xi_i^e$ and $\xi_i^I$ are the EEG embedding and image embedding obtained through encoder encoding. Although the EEG Decoder and cross attention in the figure will also participate in the training, we will not use it in subsequent tasks.}
    \label{EEG-Encoder-train}
\end{figure}

To address this limitation, we draw inspiration from the Masked AutoEncoder (MAE) framework \cite{b36,b10}, and propose a novel two-step training strategy, as shown in Fig.~\ref{EEG-Encoder-train}. This strategy incorporates both contrastive learning and cross-modal attention mechanisms within the MAE framework. Our approach not only enhances the differentiation of features across various EEG signals but also facilitates the extraction of visual semantic information embedded within those signals.

In the first step of our training strategy, we concentrate on self-supervised learning of EEG signals without incorporating visual stimulus data. Drawing inspiration from the MAE framework, we apply random masking to the input EEG signals. This is followed by a self-supervised learning process where the reconstructed outputs are compared with the original inputs. Specifically, two distinct masking patterns are applied to each EEG signal, resulting in two different reconstructions. This enables positive-contrast learning, which aims to minimize the feature distance between reconstructions originating from the same EEG signal. Conversely, negative-contrast learning maximizes the feature differences between reconstructions from different EEG signals. The objectives of self-supervised and negative-contrast learning are encoded in the following loss function: 
\vspace{-4pt} 
\begin{equation}
    \mathcal{L}_S=-\log \frac{\exp \left(s_i^{r}\cdot s_i /\tau \right)}{\exp \left(s_i^{r}\cdot s_i /\tau \right)+\sum_{j\neq i}\exp \left(s_i^{r}\cdot s_j^{r} /\tau \right)},
\end{equation}
while the positive-contrast learning objective is defined as:
\begin{equation}
    \mathcal{L}_C=-\log \frac{\exp \left(s_i^{r_1}\cdot s_i^{r_2} /\tau \right)}{\exp \left(s_i^{r_1}\cdot s_i^{r_2} /\tau \right)+\sum_{j\neq i}\exp \left(s_i^{r_1}\cdot s_j^{r_1} /\tau \right)}.
\end{equation} 
The total loss for this phase is given by:  
\begin{equation}\label{equ_rcrs}
    \mathcal{L}_1=\gamma_C \mathcal{L}_C+\gamma_S \mathcal{L}_S,
\end{equation}
where $\gamma_C$ and $\gamma_S$ are adjustable weighting parameters.

In the second step, we integrate visual stimulus data into our framework by implementing a cross-attention module. This module guides the EEG signal reconstruction process and enhances the encoder's ability to extract visual semantic information from the EEG signals. The core concept of this cross-attention mechanism is to enable bidirectional interaction between EEG and image feature embeddings. Here, the feature embedding from one modality functions as the query vector, while the embedding from the other modality acts as the key-value pair. The attention computation is represented by: 
\begin{equation}
\scalebox{0.8}{ $
\begin{aligned}
Q_I=W^{Q_I}&\xi_i^{I}+b^{Q_I}; K_E=W^{K_E}\xi_i^{e}+b^{K_E};
    V_E=W^{V_E}\xi_i^{e}+b^{V_E},\\
    &A_E(Q_I,K_E,V_E)=softmax\left( \frac{Q_I(K_E)^T}{\sqrt{d_k}} \right)V_E,
\end{aligned}$
}
\end{equation}
where $\mathbf{W}$ and $\mathbf{b}$ represent trainable weight matrices and bias terms, and \(d_k\) is the dimension of the embedding \(\xi_i^{e}\). The attention output \(A_E(Q_I, K_E, V_E)\), together with the EEG embedding \(\xi_i^{e}\), is fed into the EEG decoder to reconstruct \(e_i^r\):
\begin{equation}
    e_i^r=D_E(\xi_i^{e}+A_E(Q_I,K_E,V_E)).
\end{equation}
The image reconstruction process \(I_i^r\) follows a similar principle:
\begin{equation}
    I_i^r=D_I(\xi_i^{I}+A_I(Q_E,K_I,V_I)).
\end{equation}
Ultimately, we apply self-supervised learning once more by calculating the composite loss:
\begin{equation}\label{equ_reri}
    \mathcal{L}_2=\gamma_E(e_i-e_i^r)^2+\gamma_I(I_i-I_i^r)^2,
\end{equation}
where $\gamma_E$ and $\gamma_I$ are adjustable weighting parameters.

\subsection{3D Object Generation}\label{s3}
In the 3D generation phase, we utilize NeRF to implicitly represent 3D objects. As illustrated in Fig.~\ref{fig:model_frame}, $ \mu(\theta|e) $ signifies the NeRF distribution conditioned on EEG signals, where $ \theta $ represents NeRF parameters and $ e $ stands for various EEG signals. Building on the capability of Stable Diffusion to produce high-quality, detailed images, we employ it to guide the optimization of NeRF. However, directly applying pre-trained U-Net parameters from Diffusion and Low-Rank Adaptation (LoRA) \cite{b22} is insufficient, necessitating fine-tuning of the U-Net. The methodology for fine-tuning will be detailed in Section \ref{s4}.

In the 3D generation stage, we use the U-Net architecture within Stable Diffusion to predict the distribution of authentic images, denoted as $ p_0(x_0|e^c) $, where $ e^c $ represents the EEG signal combined with orientations determined by different camera angles, $ c $. Simultaneously, we employ a U-Net architecture optimized with LoRA, which requires minimal training data, to predict the distribution $ q_0^\mu(x_0|c,e) $ of 2D rendered images $ x_0 = g(\theta, c) $, generated by the rendering function $ g(\cdot, c) $ for varying $ c $. The objective is to align the rendered image distribution $ q_0^\mu(x_0|c, e) $ closely with the real image distribution $ p_0(x_0|e^c) $ using KL-divergence as the measure.
\\
\textbf{The Main Idea of the Update Process.}
A lower KL-divergence signifies a stronger alignment between two distributions. However, directly computing this metric is often computationally intensive. To address this, we utilize the VSD method \cite{b11} as an efficient alternative. As $ t $ approaches $ T $, as depicted in Fig.~\ref{fig:model_frame}, the optimization becomes more feasible since the diffusion distribution approaches a standard Gaussian. This leads to the formulation in Equation \ref{e16}:
\vspace{-5pt} 
\begin{equation}\label{e16}
    \mu^* := \underset{\mu}{\arg \min} \mathbb{E}_{t,c} \left[ \left( {\sigma_t}/{\alpha_t} \right) \omega(t) {D}_{KL} \left( q_t^{\mu}(\bm{x}_t | c, e)||p_t(\bm{x}_t | e^c) \right) \right].
\end{equation}
\textbf{Update Rule for $\theta$ Based on the Score Diffusion Model.}
To solve the above formula, we denote the distribution of $ \mu $ as $ \{\theta\}_{i=1}^n $ and advance optimization using the following ordinary differential equation in Equation \ref{e17}:
\begin{equation}\label{e17}
\scalebox{1}{$
\begin{aligned}
    \frac{{d\theta}_{\tau}}{d_{\tau}}=-\mathbb{E}_{t,\bm\epsilon,c}[\omega(t)\bm{(}\underbrace{-{\sigma_t}\nabla_{\bm{x}_t}\log{p_t}(\bm{x}_t|e^c)}_{\textit{score of noisy real images}}-\\\underbrace{\left(-{\sigma_t}\nabla_{\bm{x}_t}\log{q_t^{\mu_\tau}}(\bm{x}_t|c,e)\right)}_{\textit{score of noisy rendered images}}\bm{)} \frac{\partial \bm{g}(\theta_\tau,c)}{\partial \theta_\tau}].
    \end{aligned}$
}
\end{equation}
Solutions to this challenge are found through score-based diffusion models, which allow for the gradual optimization of $ \theta $. The score function acts as an approximation of the gradient. For real images, the score function, represented as $ -\sigma_t\nabla_{x_t}\log p_t(x_t|e^c) $, is derived from the Stable Diffusion model $ \epsilon_{\textup{pretrain}}(x_t, t, e^c) $. Conversely, the score function for rendered images, $ -\sigma_t\nabla_{x_t}\log q_t^{\mu_\tau}(x_t|c, e) $, is calculated using LoRA $ \epsilon_\phi(x_t, t, c, e) $. It is important to note that LoRA requires prior training with rendered images before each noise prediction.

In summary, the optimization process is expressed as $ \theta^{(i)} \leftarrow \theta^{(i)} - \eta \nabla_{\theta}\mathcal{L}_{\textup{VSD}}(\theta^{(i)}) $, where $ \eta $ is the learning rate. Based on Equation \ref{e17}, the gradient expression, Equation \ref{e18}, is:
\begin{equation}\label{e18}
\scalebox{0.75}{$
\begin{aligned}
 {\nabla_{\bm{\theta}}}\mathcal{L}_{\textup{VSD}}(\theta) \overset{\triangle}{=}\mathbb{E}_{t,\bm\epsilon,c}\left[ \omega(t)({\bm\epsilon}_{pretrain}(\bm{x}_t,t,e^c))-{\bm\epsilon}_{\phi}(\bm{x}_t,t,c,e)\frac{\partial \bm{g}(\theta,c)}{\partial \theta}\right],
    \end{aligned}$
}
\end{equation}
where $x_t=\alpha_tg(\theta,c)+\sigma_t\epsilon$.

    

\section{Experiment}

\subsection{Dataset and Data Processing}
EEG records the electrical activity of the human brain. For our study, we utilized a publicly available dataset from the MOBBA platform \cite{b15}, which consists of EEG-ImageNet pairs. This dataset includes approximately 40 categories, encompassing around 2000 images with corresponding EEG recordings from subjects. During the EEG recording sessions, each image was shown for 0.5 seconds, followed by a 10-second break after every 50 images. EEG data were collected using a set of 128 electrodes placed on the scalp of each participant.

In terms of data preprocessing, we initially filtered all EEG signals to a frequency range of 5-95 Hz and truncated the signal length to 512 time points. Due to the low Signal-to-Noise Ratio and occasional gaps in the EEG data, the signals were padded to maintain 128 channels. Subsequently, every four consecutive time steps were combined into a single token, which was then transformed into a 1024-dimensional embedding using a projection layer for further masked signal modeling.

\subsection{Deatils of Finetuning the U-Net in Diffusion and LoRA}\label{s4}

Fine-tuning, while seemingly straightforward, is crucial for enhancing the quality and accuracy of the model's outputs. It is only when Stable Diffusion accurately predicts the distribution of real images, denoted as \( p_0(x_0|e^c) \), and LoRA precisely estimates the distribution of rendered images, \( q_0^\mu(x_0|c, e) \), that we can effectively guide the creation of high-quality 3D effects. Specifically, we fine-tune the U-Net structure within both Stable Diffusion and LoRA using a limited dataset of EEG-image pairs. This fine-tuning is integrated into the Stable Diffusion framework.

The conditional signal is applied through a cross-attention mechanism within the U-Net architecture. We use the output vector \( \xi_i^{e} \) from the EEG encoder as the controlling condition for the U-Net in both Stable Diffusion and LoRA. Additionally, we employ time embeddings as supplementary control conditions. Our objective is to enable Stable Diffusion and LoRA to accurately predict noise with EEG embeddings as controlling conditions.

During the fine-tuning process, we keep the parameters of the EEG encoder and all components of the Stable Diffusion model fixed, except for those within the U-Net architecture. Only the cross-attention heads in the U-Net are trained. We incorporate the EEG encoder's output \( \xi_i^{e} \) and the image \( I \) from the EEG-image pairs into the \( Q, K, \) and \( V \) components of the cross-attention module within the U-Net, respectively. This integration ensures the EEG embedding is effectively incorporated into the U-Net architecture through the cross-attention layer.

During implementation, based on EEG-image pairs, we fine-tune the U-Net using the following loss function:
\vspace{-4pt} 
\begin{equation}
L_{SD} = \mathbb{E}_{I, \epsilon \sim \mathcal{N}(0,1), t}\left[\| \epsilon - \epsilon_\theta\left(I_t, t, \tau_\theta({\xi_i^{e}})\right) \|_2^2\right],
\end{equation}
where $\epsilon_\theta$ represents the denoising function implemented by U-Net.


\subsection{Implementation Details}
Our model was trained using NVIDIA A100 GPUs, and we detail the parameter settings for each training phase below. In Stage 1 of the MERL process, we set the masking rate for EEG signals at 0.75, a configuration that continues into Stage 2. The model uses an asymmetric EEG auto-encoder architecture, with the encoder comprising 24 layers and the decoder 8 layers, and rate parameters set to \( \gamma_C = 0.5 \) and \( \gamma_S = 1 \). We used a batch size of 10, as contrastive learning requires a batch size greater than 1. The training spanned 150 epochs with a learning rate of \( 2.5 \times 10^{-4} \). Optimization was performed using the AdamW \cite{b27} optimizer with a weight decay of 0.05.

In Stage 2, we adjusted both the EEG auto-encoder and the image auto-encoder. For the image auto-encoder, the encoder has 12 layers and the decoder has 6 layers. We kept the EEG masking rate at 0.75 and set the image masking rate to 0.5, with \( \gamma_E = 0.25 \) and \( \gamma_I = 1.5 \). The batch size was set to 4, with 60 epochs and a learning rate of \( 5.3 \times 10^{-5} \). The AdamW optimizer was again used with a weight decay of 0.05.

Throughout the training, we chose to freeze the parameters of the decoder and focused the training exclusively on the encoder component of the auto-encoder, as the decoder will be discarded in later stages of our methodology.

\textbf{3D Generation.} In this phase, we concentrate on producing multi-view 2D images and segmenting only the central object to build 3D models. This method is based on the finding that including the background in 3D video generation does not lead to satisfactory results. For this process, we set the Classifier-Free Guidance (CFG) to 3 and the particle number for VSD to 4. The learning rate is established at \( 1 \times 10^{-3} \), with a batch size of 4, an epoch limit of 1000, and we use the AdamW optimizer for training.

\textbf{Finetune U-Net.} We refine the U-Net architecture using a limited dataset of EEG-image pairs through a Stable Diffusion-based approach. The fine-tuning parameters include 50 epochs, a batch size of 8, 1000 diffusion steps, the utilization of the AdamW optimizer, a learning rate of $ 1 \times 10^{-4} $, and an image resolution of $512\times512\times3$.

\subsection{Comparisons Experiment}
\begin{table*}[htbp]
\caption{Ablation experiments on training strategies. The best results are shown in bold.}
\label{tab:my_table1}
\centering
\tiny
\resizebox{0.95\textwidth}{!}{
\begin{tabular}{c|c|c|c|c|c|c}
\Xhline{1px}
\textbf{\makecell{Self-supervised\\and Negative\\Contrast Learning}} & \textbf{\makecell{Positive\\Contrast Learning}} & \textbf{\makecell{EEG Cross-Attention\\Self-supervised\\Learning}} & \textbf{\makecell{Image Cross-Attention\\Self-supervised\\Learning}} & \textbf{PSNR$\uparrow$} & \textbf{LPIPS$\downarrow$} & \textbf{FID$\downarrow$} \\
\hline\hline
$\checkmark$ & $\checkmark$ & $\checkmark$ &  & 24.165 & 0.532 & 140.225 \\
$\checkmark$ & $\checkmark$ &  & $\checkmark$ & 22.356 & 0.694 & 177.414 \\
$\checkmark$ &  & $\checkmark$ & $\checkmark$ & 21.524 & 0.722 & 195.462 \\
 & $\checkmark$ & $\checkmark$ & $\checkmark$ & 22.014 & 0.702 & 181.424 \\
$\checkmark$ & $\checkmark$ & $\checkmark$ & $\checkmark$ & \textbf{28.520} & \textbf{0.318} & \textbf{57.333} \\
\Xhline{1px}
\end{tabular}
}
\end{table*}
3D-Telepathy is pioneering in generating three-dimensional representations from EEG signals. To date, only two studies have explored EEG-to-3D generation \cite{b44, b45}, but neither has provided access to their code. To assess the performance of 3D-Telepathy, we implemented a comprehensive, multi-step evaluation protocol. Since NeRF produces 3D video sequences while the reference data in our study consists of static images, we first extracted frames from the NeRF-generated videos. This conversion of continuous 3D renderings into discrete image sequences enabled a quantitative evaluation of rendering results from specific viewpoints. Next, we established a rigorous viewpoint-matching mechanism by aligning the generated frames with the viewpoints of images from the EEG-ImageNet dataset. Additionally, we standardized image resolution and dimensions to create a reliable foundation for quantitative analysis. This process ensures fair comparisons with existing methods for EEG-to-2D image reconstruction. For evaluation, we employed three complementary metrics: (1) PSNR, to assess pixel-level reconstruction quality; (2) FID, to measure distributional similarity between generated and reference images; and (3) LPIPS, to evaluate perceptual similarity. The comparative results for these metrics are detailed in Table \ref{tab:Comparison_experiment_table}.

\begin{table}[htbp]
    \scriptsize
    \centering
    \caption{Performance Comparisons of the images generated by each model.}
    \label{tab:Comparison_experiment_table}
    \resizebox{0.45\textwidth}{!}{
    \begin{tabular}{c|c|c|c}
    \specialrule{.2em}{0em}{.2em}
    & PSNR $\uparrow$ & LPIPS $\downarrow$ & FID $\downarrow$ \\
    \hline
    CapsEEGNet \cite{b28} & 28.0485 & 0.3269 & 64.0343 \\
    Brain2Image \cite{b3} & 25.9264 & 0.5735 & 129.8699 \\
    DreamDiffusion \cite{b2} & 27.6997 & 0.3774 & 109.6240 \\
    Dongyang Li \textit{et al.} \cite{b29} & 27.8463 & 0.3287 & 63.0432 \\
    \textbf{3D-Telepathy (Ours)} & \textbf{28.5203} & \textbf{0.3178} & \textbf{57.3326} \\
    \specialrule{.2em}{0em}{.2em}
    \end{tabular}
    }
\end{table}

\noindent\textbf{CapsEEGNet \cite{b28}} introduces an innovative EEG encoder that leverages primary capsules, digital capsules, and transformer layers to capture EEG information. \ 

\noindent\textbf{Brain2Image \cite{b3}} utilizes a Long Short-Term Memory (LSTM) \cite{b34} neural network as the EEG encoder and a GAN \cite{b35} for image generation.\ 

\noindent\textbf{DreamDiffusion \cite{b2}} trains its EEG encoder using a dual contrastive masking approach. The EEG embeddings are then introduced into LDM through a cross-attention module to generate final images.

\noindent\textbf{Dongyang Li \textit{et al.} \cite{b29}} This article proposes an EEG-based zero-sample visual decoding and reconstruction framework with an innovative design of an EEG encoder called Adaptive Thinking Mapper (ATM).\

Our method compared to Brain2Image, CapsEEGNet added the Stable Diffusion module with strong image generation and cross-modal generation capabilities, so therefore PSNR went up by 2.5939 and 0.4718, LPIPS went down by 0.2557 and 0.0091, and FID went down by 72.5373 and 6.7017, respectively. Compared to DreamDiffusion and Dongyang Li \textit{et al.} our EEG encoder was trained using self-contrast combined with cross-contrast and tuned with cross-attention, so as a result the PSNR went up by 0.8206 and 0.6740, the LPIPS went down by 0.0596 and 0.0109, and FID decreased by 52.2914 and 5.7097, respectively.

Each of these models offers unique approaches to the technical challenge of reconstructing visual data from EEG signals. 3D-Telepathy stands out not only for its breakthrough in reconstructing 3D visual objects from EEG signals, but also for its dual self-attention mechanism in the EEG encoder. Integrates cross-attention and contrast learning within a masked self-supervised training framework, leading to more effective feature extraction and learning.

\subsection{Ablation Study for Training Strategy}

\begin{figure*}[htbp]
    \centering
    \includegraphics[width=0.9\linewidth]{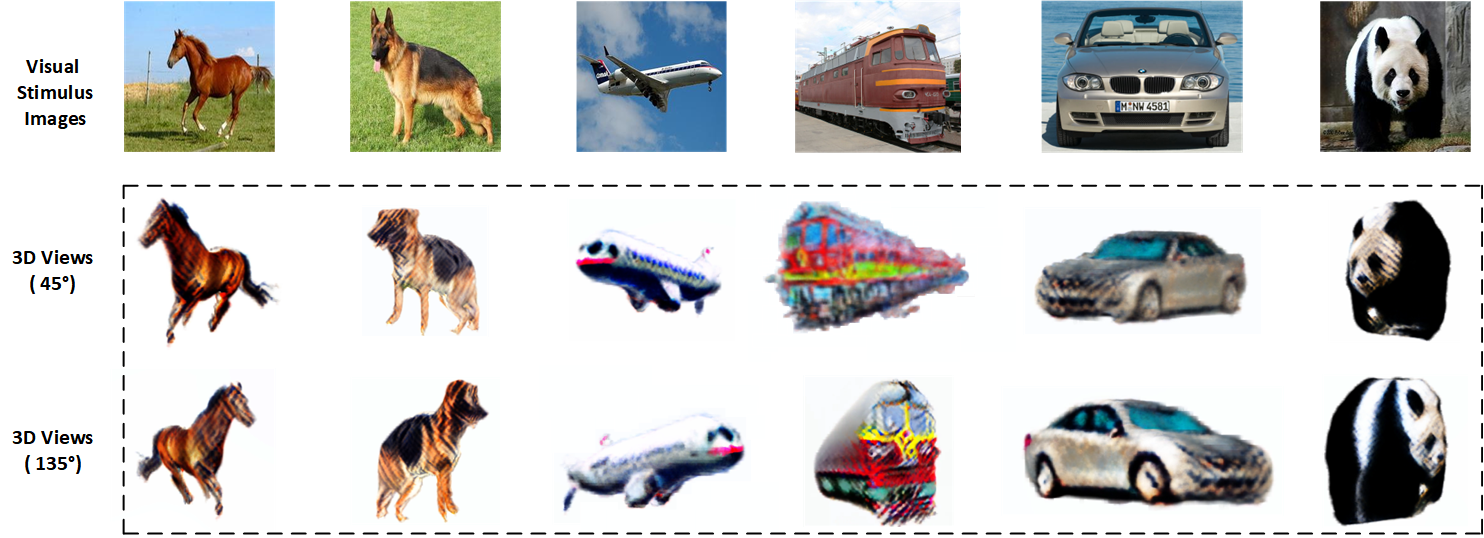}
    \caption{The results of 3D reconstruction of an EEG signal using 3D-Telepathy are shown here for $45^{\circ}$ and $135^{\circ}$, respectively, and the corresponding visual stimuli for this EEG signal. }
    \label{r1}
\end{figure*}

To thoroughly assess the effectiveness of our proposed method, we analyzed the EEG encoder training strategy by varying the weighting coefficients in the loss function and conducting ablation experiments for each training approach. We used the same methodology used in our ablation experiments. This involved extracting frames from NeRF-generated videos and aligning them precisely with the viewpoints of images from the EEG-ImageNet dataset. Table \ref{tab:my_table1} displays the results of these experiments. Columns 1 and 2 show the training methods used in Fig.~\ref{EEG-Encoder-train} Step 1. Similarly, the data in the 3 and 4 columns show the necessity of EEG and Image Cross-Attention Self-supervised Learning in Step 2 of Fig.~\ref{EEG-Encoder-train}. 

The experimental results indicate that omitting any training approach results in a significant decline in final output quality. Experimental results demonstrate that when removing Positive Contrast Learning, Self-supervised and Negative Contrast Learning from Fig.~\ref{EEG-Encoder-train} Step 1, the model performance significantly deteriorates: PSNR decreases by 23.7\%, LPIPS increases by 123.9\%, and FID value rises by 228.7\%. Similarly, eliminating EEG Cross-Attention Self-supervised Learning and Image Cross-Attention Self-supervised Learning from Fig.~\ref{EEG-Encoder-train} Step 2 results in a 18.4\% decrease in PSNR, a 92.8\% increase in LPIPS, and a 177.0\% rise in FID value. It reveals that while all training strategies contribute positively to generation quality, the three training strategies in Step 1 demonstrate notably greater importance. This phenomenon may be attributed to the fact that this stage enables the EEG encoder to develop effective clustering capabilities for different EEG signals.

As shown in Fig.~\ref{r1}, we perform a multi-view comparative analysis between our EEG-reconstructed 3D objects and the original 2D images. The experimental results demonstrate that our method effectively reconstructs the visible sides of objects from 2D images into their corresponding 3D forms. However, for object sides not captured in the original viewpoints, there are notable limitations in maintaining structural and color fidelity in the 3D reconstruction. These findings suggest that while our approach successfully extracts and transforms visual information encoded in EEG signals into 3D representations, it currently does not possess the capability to intelligently fill in missing visual information absent from the original EEG signals.

\subsection{Influence of Hyper-parameters}

\begin{table*}[htbp]
\caption{Experiments on the effect of hyper-parameters and fine-tuning on results. The best results are shown in bold.}
\label{tab:my_table2}
\centering
\tiny

\resizebox{0.95\textwidth}{!}{
\begin{tabular}{c|c|c|c|c|c|c|c|c|c}
\Xhline{1px}
\textbf{$\gamma_{S}$} & \textbf{$\gamma_{C}$} & \textbf{$\gamma_{E}$} & \textbf{$\gamma_{I}$} & \textbf{EEG Mask Ratio} & \textbf{Image Mask Ratio} & \textbf{Finetune} & \textbf{PSNR$\uparrow$} & \textbf{LPIPS$\downarrow$} & \textbf{FID$\downarrow$} \\
\hline\hline
0.5 & 1 & 0.5 & 0.5 & 0.5 & 0.5 & 1 & 25.473 & 0.496 & 122.140 \\
1 & 0.5 & 0.5 & 0.5 & 0.5 & 0.5 & 1 & 27.012 & 0.425 & 105.597 \\
1 & 0.5 & 0.25 & 1.5 & 0.5 & 0.5 & 1 & 27.865 & 0.479 & 76.647 \\
1 & 0.5 & 1.5 & 0.25 & 0.5 & 0.5 & 1 & 26.167 & 0.462 & 112.824 \\
1 & 0.5 & 0.25 & 1.5 & 0.5 & 0.75 & 1 & 27.641 & 0.414 & 93.252 \\
1 & 0.5 & 0.25 & 1.5 & 0.75 & 0.5 & 0 & 23.942 & 0.646 & 151.424 \\
\textbf{1} & \textbf{0.5} & \textbf{0.25} & \textbf{1.5} & \textbf{0.75} & \textbf{0.5} & \textbf{1} & \textbf{28.520} & \textbf{0.318} & \textbf{57.333} \\
\Xhline{1px}
\end{tabular}
}
\end{table*}
In this section, we explore the impact of hyperparameters on the performance of the 3D-Telepathy model, as detailed in Table \ref{tab:my_table2}. Our experiments indicate that the weighting parameters \(\gamma_S\) and \(\gamma_C\) significantly influence the performance of the model during the encoder training phase. The model reaches optimal performance when \(\gamma_S = 1\) and \(\gamma_C = 0.5\). Further testing on the re-parameterization of \(\gamma_E\) and \(\gamma_I\) reveals that the best results occur when \(\gamma_E = 0.25\) and \(\gamma_I = 1.5\), highlighting the importance of integrating image data and multimodal learning for extracting visual information from EEG signals. 

Additionally, the fine-tuning process is essential for enhancing model performance, and experiments demonstrate a significant improvement in output quality through fine-tuning. In conclusion, after thorough experimentation, we have identified a set of optimal parameter configurations and methodological choices, as summarized in the last row of Table \ref{tab:my_table2}.

\begin{figure}[htbp]
    \centering
    \includegraphics[width=1\linewidth]{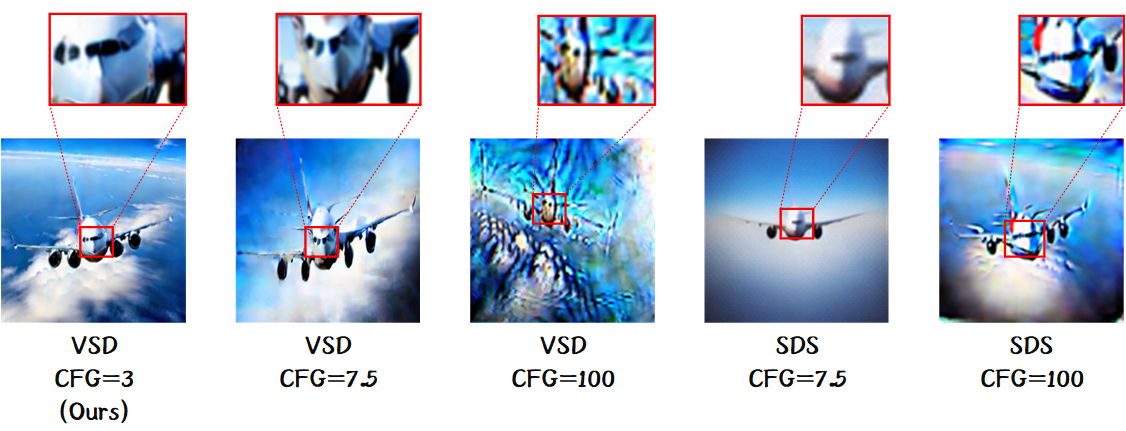}
    \caption{Different results obtained using VSD or SDS methods and different CFG. After a plenty of experiments about the generated images, we determined that VSD, CFG=3 is the best combination.}
    
    \label{VSD SDS}
\end{figure}

The masking rates for EEG and images are also crucial, set at 0.75 and 0.5, respectively, to effectively optimize EEG signal processing. In NeRF training, we compared VSD and SDS methods, finding that VSD is more effective for generating high-quality 3D objects. The CFG value plays a critical role in maintaining the semantic consistency of the generated results, and our experiments show that CFG = 3 is the most effective choice. The results of using the VSD or SDS methods and different CFG values are illustrated in Fig.~\ref{VSD SDS}.


\section{Conclusion}
In this study, we introduce a groundbreaking task focused on reconstructing 3D objects from EEG signals. Our model architecture is divided into two primary stages. The first stage involves an EEG encoder equipped with a dual self-attention mechanism, which extracts visual stimulus information from EEG signals. In the second stage, the EEG embeddings are mapped to a U-Net architecture to reconstruct 3D objects through NeRF training using the VSD method. To enhance the effectiveness of the EEG encoder, we employ a comprehensive training strategy that integrates cross-attention, dual-masked contrastive learning, and self-supervised learning.

This task poses significant challenges due to the inherent noise in EEG signals and the complex nature of deriving 3D reconstructions from 2D observations. These challenges contribute to issues such as the loss of fine texture details in the generated 3D objects. Although our current results show limitations in texture detail preservation and structural accuracy, this work pioneers a new direction at the intersection of 3D generation and neuro-imaging, pushing the boundaries of cross-modal 3D generation.

\section{Acknowledgments}
This work was supported by the Open Project Program of the State Key Laboratory of CAD\&CG [Grant No. A2410], Zhejiang University; Zhejiang Provincial Natural Science Foundation of China [No. LY21F020017, 2023C03090]; National Natural Science Foundation of China [No. 61702146, 62076084, U20A20386, U22A2033]; Guangdong Basic and Applied Basic Research Foundation [No. 2025A1515011617, 2022A1515110570]; Shenzhen Longgang District Science and Technology Innovation Special Fund [No. LGKCYLWS2023018] and Shenzhen Medical Research Fund [No. C2401036].

\vspace{12pt}


\begin{thebibliography}{00}
\bibitem{b1}
A.~E. Welchman, ``The human brain in depth: how we see in 3d,'' {\em Annual Review of Vision Science}, vol.~2, no.~1, pp.~345--376, 2016.

\bibitem{b2} Y.~Bai, X.~Wang, Y.~Cao, Y.~Ge, C.~Yuan, and Y.~Shan, ``Dreamdiffusion: Generating high-quality images from brain eeg signals,'' {\em arXiv preprint arXiv:2306.16934}, 2023.


\bibitem{b3} I.~Kavasidis, S.~Palazzo, C.~Spampinato, D.~Giordano, and M.~Shah, ``Brain2image: Converting brain signals into images,'' in {\em Proceedings of the 25th ACM International Conference on Multimedia}, MM '17, (New York, NY, USA), p.~1809–1817, Association for Computing Machinery, 2017.





\bibitem{b5} R.~Rombach, A.~Blattmann, D.~Lorenz, P.~Esser, and B.~Ommer, ``High-resolution image synthesis with latent diffusion models,'' in {\em 2022 IEEE/CVF Conference on Computer Vision and Pattern Recognition (CVPR)}, pp.~10674--10685, 2022.



\bibitem{b8} Y.~Takagi and S.~Nishimoto, ``High-resolution image reconstruction with latent diffusion models from human brain activity,'' in {\em 2023 IEEE/CVF Conference on Computer Vision and Pattern Recognition (CVPR)}, pp.~14453--14463, 2023.



\bibitem{b9} Y.~Lu, C.~Du, Q.~Zhou, D.~Wang, and H.~He, ``Minddiffuser: Controlled image reconstruction from human brain activity with semantic and structural diffusion,'' in {\em Proceedings of the 31st ACM International Conference on Multimedia}, MM '23, (New York, NY, USA), p.~5899–5908, Association for Computing Machinery, 2023.




\bibitem{b10} J.~Sun, M.~Li, Z.~Chen, Y.~Zhang, S.~Wang, and M.-F. Moens, ``Contrast, attend and diffuse to decode high-resolution images from brain activities,'' in {\em Advances in Neural Information Processing Systems} (A.~Oh, T.~Naumann, A.~Globerson, K.~Saenko, M.~Hardt, and S.~Levine, eds.), vol.~36, pp.~12332--12348, Curran Associates, Inc., 2023.

\bibitem{b11} Z.~Wang, C.~Lu, Y.~Wang, F.~Bao, C.~LI, H.~Su, and J.~Zhu, ``Prolificdreamer: High-fidelity and diverse text-to-3d generation with variational score distillation,'' in {\em Advances in Neural Information Processing Systems} (A.~Oh, T.~Naumann, A.~Globerson, K.~Saenko, M.~Hardt, and S.~Levine, eds.), vol.~36, pp.~8406--8441, Curran Associates, Inc., 2023.


\bibitem{b12} B.~Poole, A.~Jain, J.~T. Barron, and B.~Mildenhall, ``Dreamfusion: Text-to-3d using 2d diffusion,'' {\em arXiv preprint arXiv:2209.14988}, 2022.


\bibitem{b13} J.~Gao, Y.~Fu, Y.~Wang, X.~Qian, J.~Feng, and Y.~Fu, ``Mind-3d: Reconstruct high-quality 3d objects in human brain,'' in {\em European Conference on Computer Vision}, pp.~312--329, Springer, 2024.

\bibitem{b14} B.~Mildenhall, P.~P. Srinivasan, M.~Tancik, J.~T. Barron, R.~Ramamoorthi, and R.~Ng, ``Nerf: representing scenes as neural radiance fields for view synthesis,'' {\em Commun. ACM}, vol.~65, p.~99–106, Dec. 2021.


\bibitem{b15} V.~Jayaram and A.~Barachant, ``MOABB: trustworthy algorithm benchmarking for bcis,'' {\em Journal of Neural Engineering}, vol.~15, p.~066011, Sept. 2018.


\bibitem{b22} E.~J. Hu, Y.~Shen, P.~Wallis, Z.~Allen-Zhu, Y.~Li, S.~Wang, L.~Wang, and W.~Chen, ``Lo{RA}: Low-rank adaptation of large language models,'' in {\em International Conference on Learning Representations}, 2022.




\bibitem{b27} I.~Loshchilov and F.~Hutter, ``Decoupled weight decay regularization,'' {\em arXiv preprint arXiv:1711.05101}, 2019.

\bibitem{b28} X.~Deng, Z.~Wang, K.~Liu, and X.~Xiang, ``A gan model encoded by capseegnet for visual eeg encoding and image reproduction,'' {\em Journal of Neuroscience Methods}, vol.~384, p.~109747, 2023.



\bibitem{b29} D.~Li, C.~Wei, S.~Li, J.~Zou, and Q.~Liu, ``Visual decoding and reconstruction via eeg embeddings with guided diffusion,'' in {\em Advances in Neural Information Processing Systems} (A.~Globerson, L.~Mackey, D.~Belgrave, A.~Fan, U.~Paquet, J.~Tomczak, and C.~Zhang, eds.), vol.~37, pp.~102822--102864, Curran Associates, Inc., 2024.


\bibitem{b34} K.~Greff, R.~K. Srivastava, J.~Koutník, B.~R. Steunebrink, and J.~Schmidhuber, ``Lstm: A search space odyssey,'' {\em IEEE Transactions on Neural Networks and Learning Systems}, vol.~28, no.~10, pp.~2222--2232, 2017.



\bibitem{b35} I.~Goodfellow, J.~Pouget-Abadie, M.~Mirza, B.~Xu, D.~Warde-Farley, S.~Ozair, A.~Courville, and Y.~Bengio, ``Generative adversarial networks,'' {\em Commun. ACM}, vol.~63, p.~139–144, Oct. 2020.



\bibitem{b36} K.~He, X.~Chen, S.~Xie, Y.~Li, P.~Dollár, and R.~Girshick, ``Masked autoencoders are scalable vision learners,'' in {\em 2022 IEEE/CVF Conference on Computer Vision and Pattern Recognition (CVPR)}, pp.~15979--15988, 2022.



\bibitem{b38} Y.~Song, B.~Liu, X.~Li, N.~Shi, Y.~Wang, and X.~Gao, ``Decoding {{Natural Images}} from {{EEG}} for {{Object Recognition}},'' in {\em International {{Conference}} on {{Learning Representations}}}, 2024.



\bibitem{b39}C.~Meng, Y.~He, Y.~Song, J.~Song, J.~Wu, J.-Y. Zhu, and S.~Ermon, ``{SDE}dit: Guided image synthesis and editing with stochastic differential equations,'' in {\em International Conference on Learning Representations}, 2022.

\bibitem{b40}R.~Liu, R.~Wu, B.~V. Hoorick, P.~Tokmakov, S.~Zakharov, and C.~Vondrick, ``Zero-1-to-3: Zero-shot one image to 3d object,'' {\em arXiv preprint arXiv:2303.11328}, 2023.









\bibitem{b44}Z.~Guo, J.~Wu, Y.~Song, J.~Bu, W.~Mai, Q.~Zheng, W.~Ouyang, and C.~Song, ``Neuro-3d: Towards 3d visual decoding from eeg signals,'' {\em arXiv preprint arXiv:2411.12248} 2024.


\bibitem{b45}X.~Xiang, W.~Zhou, and G.~Dai, ``EEG-driven 3d object reconstruction with style consistency and diffusion prior,'' {\em arXiv preprint arXiv:2410.20981} 2024.


\end{thebibliography}
\end{document}